\documentclass[graybox]{svmult}
\usepackage{mathptmx}      
\usepackage{helvet}         
\usepackage{courier}        
\usepackage{type1cm}       
\usepackage{makeidx}         
\usepackage{amsmath,amsfonts,amssymb}
\usepackage{graphicx}       
\usepackage{multicol}        
\usepackage[bottom]{footmisc}
\usepackage{subcaption}
\usepackage{placeins}
\usepackage{algorithm}
\usepackage{algpseudocode}
\usepackage{tikz}
\usepackage{tikz-3dplot}
\usepackage{float}
\usepackage{titlesec}
\usepackage[a4paper,
    left=2.50cm,
    right=2.50cm,
    top=2.70cm,
    bottom=2.70cm
]{geometry}

\titlespacing*{\subsection}{0pt}{10pt}{3pt}
\titlespacing*{\section}{0pt}{20pt}{3pt}
\titlespacing*{\subsubsection}{0pt}{10pt}{3pt}
\usetikzlibrary{arrows.meta,calc,decorations.pathreplacing}

\makeindex             

\begin{document}

\title*{Mobility-Aware Mobile Edge Computing for Energy-Balanced Data Collection in IoT-Enabled Underwater Sensor Networks}
\titlerunning{Mobility-Aware MECg for Energy-Balanced Data Collection in IoT-Enabled UWSNs} 
\author{Thi My Binh Nguyen*, Dinh Hoan Do, Tien Manh Tong, Minh Yen Vu, Trong Hop Dang, and Luong Ho}
\authorrunning{Binh et. al.} 
\institute{Nguyen Thi My Binh \at Hanoi University of Industry, Hanoi, Vietnam \email{binhntm@haui.edu.vn}
\and Do Dinh Hoan \at Hanoi University of Science and Technology, Hanoi, Vietnam \email{hoan.dd224986@sis.hust.edu.vn}
\and Tong Tien Manh \at Hanoi University of Science and Technology, Hanoi, Vietnam \email{manh.tt225040@sis.hust.edu.vn}
\and Vu Minh Yen \at Hanoi University of Industry, Hanoi, Vietnam \email{yenvm@haui.edu.vn}
\and Dang Trong Hop \at Hanoi University of Industry, Hanoi, Vietnam \email{hopdt@haui.edu.vn}
\and Ho Viet Duc Luong \at Hanoi University of Science and Technology, Hanoi, Vietnam \email{luonghvd@soict.hust.edu.vn}}
\maketitle

\abstract{Underwater wireless sensor networks (UWSNs) face critical challenges in energy imbalance, connectivity, and data collection due to harsh  environments. To address these issues, this paper focuses on the problem is how to achieve energy-balanced data collection in IoT-enabled UWSNs with a mobility-aware mobile edge computing (MEC), named MECEBDC. Unlike existing approaches, the proposed scheme jointly optimizes energy-aware clustering, cluster head selection, and AUV-assisted data collection. 
Furthermore, we design a novel metaheuristic-based AUV routing algorithm that incorporates realistic underwater mobility dynamics, explicitly considering gravitational pressure and buoyancy forces acting on the AUV. By leveraging the computing, storage, and mobility capabilities of MEC-enabled AUVs, the proposed approach enables efficient trajectory planning and adaptive data gathering in a stable 3D underwater environment. Extensive simulations and theoretical analysis demonstrate that the proposed method significantly reduces node energy consumption, improves data collection efficiency, and prolongs network lifetime compared with existing schemes, making it well suited for long-term underwater IoT applications.}

\section{Introduction}
\label{sec1}
The rapid expansion of the Internet of Things (IoT) has enabled intelligent data acquisition in diverse domains such as environmental monitoring and industrial automation \cite{asghari2019internet}. Extending IoT to underwater environments has led to the emergence of underwater wireless sensor networks (UWSNs), which support large-scale and long-term ocean observation \cite{gupta2024underwater}. In UWSNs, sensor nodes are typically deployed in a three-dimensional underwater space and collaboratively collect data for transmission to surface base stations \cite{sandeep2017review}. However, harsh underwater conditions including limited bandwidth, high propagation delay, dynamic topology, and strict energy constraints—pose significant challenges to efficient and sustainable data collection.
Traditional UWSN data collection schemes primarily rely on multi-hop acoustic communication, where data are forwarded through intermediate nodes \cite{fanian2019cluster}. Although simple, this approach often causes severe energy imbalance, particularly for nodes near the base station, resulting in premature node failures and reduced network lifetime. To mitigate these issues, autonomous underwater vehicles (AUVs) have been introduced as mobile data collectors, enabling short-range communication and alleviating energy hotspots. Nevertheless, many existing AUV-assisted schemes suffer from high latency and suboptimal performance due to oversimplified mobility assumptions and inefficient trajectory planning.

MEC has recently emerged as a promising paradigm for enhancing data processing efficiency by bringing computing and storage resources closer to data sources \cite{mao2017survey,haibeh2022survey}. In underwater IoT scenarios, AUVs naturally serve as mobile edge nodes thanks to their onboard computation capability, storage resources, and controllable mobility. However, current MEC-enabled UWSN solutions often fail to fully exploit these advantages. In particular, realistic underwater mobility dynamics, such as buoyancy, gravitational pressure, and three-dimensional motion—are commonly neglected, while clustering and cluster head selection are rarely optimized jointly for energy balance and network connectivity.
To address these limitations, this paper proposes a mobility-aware MEC-based framework for energy-balanced data collection in IoT-enabled UWSNs, referred to as MECEBDC. The proposed scheme integrates energy- and connectivity-aware clustering, intelligent cluster head selection, and AUV-assisted data gathering into a unified design. Moreover, a novel metaheuristic-based AUV routing algorithm is developed that explicitly incorporates realistic underwater mobility forces, enabling adaptive and efficient trajectory planning in a stable three-dimensional environment. Extensive simulations and theoretical analysis demonstrate that the proposed framework significantly reduces energy consumption, improves data collection efficiency, and prolongs network lifetime, making it well suited for long-term underwater IoT applications.

\section{Related works}\label{sec:related-works}
UWSNs have become a key technology for underwater monitoring and Internet of Underwater Things applications. However, the underwater environment poses severe challenges, including limited bandwidth, high propagation delay, dynamic topology, and strict energy constraints, which significantly degrade data collection efficiency and network lifetime \cite{gupta2024underwater}. As a result, extensive research efforts have focused on energy-efficient routing, clustering, and AUV-assisted data collection mechanisms.
Early studies mainly addressed energy efficiency through routing protocol design. Nguyen et al. proposed an energy-efficient and localization-free routing protocol to reduce delay and improve packet delivery while maintaining acceptable energy consumption \cite{nguyen2025innovative}. Although routing-based approaches enhance connectivity and transmission reliability, multi-hop communication still leads to uneven energy consumption and rapid energy depletion of relay nodes in large-scale UWSNs.

To overcome the limitations of multi-hop transmission, AUV-assisted data collection has been widely investigated. Nam et al. developed an enhanced lawn mower pattern path (ELMPP) to support long-range AUV operations and energy consumption among nodes \cite{nam2018data}. Nevertheless, this work does not explicitly consider AUV maneuvering costs and realistic underwater mobility dynamics, which may significantly impact mission duration. Similarly, Zhuo et al. formulated clustering, MAC coordination, and AUV path planning as optimization problems to balance energy consumption and throughput \cite{zhuo2020auv}. However, AUV trajectory planning was simplified as a TSP, ignoring physical forces acting on the AUV.
More recently, MEC has been introduced into UWSNs to improve data processing and collection efficiency. Cai et al. proposed a MEC-enabled AUV-based data collection scheme that considers AUV velocity and direction in a stable 3D underwater environment \cite{cai2019data}. While this approach improves energy efficiency and network lifetime, clustering and cluster head selection are not jointly optimized in an energy-aware manner. Other studies, such as DCCHS \cite{tang2025enhancing} and ACRL \cite{chen2025auv}, incorporate dynamic clustering, residual energy-based cluster head selection, and intelligent AUV trajectory planning using optimization or reinforcement learning techniques. These methods effectively balance energy consumption; however, they typically rely on simplified AUV mobility models and do not explicitly account for underwater physical forces such as gravitational pressure and buoyancy.

In parallel, survey studies have highlighted the importance of clustering and metaheuristic-based optimization in controlling energy consumption in WSNs \cite{fanian2019cluster}. Despite their potential, the integration of metaheuristic optimization with MEC-enabled AUVs and realistic underwater mobility modeling remains limited.
In summary, existing works have made notable progress in routing, clustering, and AUV-assisted data collection for UWSNs. However, few studies jointly optimize energy-aware clustering, reliable cluster head selection, and AUV trajectory planning while explicitly considering realistic underwater mobility dynamics within a MEC framework. These gaps motivate the proposed MECEBDC scheme, which leverages MEC-enabled AUVs and a physics-aware mobility model to achieve energy-balanced and efficient data collection in IoT-enabled UWSNs.

\section{Preliminaries and problem formulation}\label{sec:preliminaries-problem}
In IoT-enabled UWSNs, the performance of data
acquisition is mainly evaluated in terms of network energy consumption and data
collection latency. We first presents the energy consumption model, the data
collection delay model, and the mobility model of the AUV.

\textbf{Network energy consumption model.}
At each data collection cycle $t$, the total energy consumption of the network
consists of the energy consumed for data collection and data transmission, which
is expressed as
\begin{equation}
E_{\text{total}}(t) = E_{\text{coll}}(t) + E_{\text{trans}}(t).
\end{equation}
The energy required to collect a data packet of length $l$ is given by
\begin{equation}
E_{\text{coll}}(t) = h\, P_c(t)\, T_c(t),
\end{equation}
where $P_c(t)$ and $T_c(t)$ denote the data collection power and duration,
respectively, and $h$ is a scaling coefficient.
Following~\cite{gupta2024underwater}, the energy consumption for data transmission is modeled as
\begin{equation}
E_{\text{trans}}(t) = P_t(t)\, T_t(t)\, A(d),
\end{equation}
where $P_t(t)$ and $T_t(t)$ denote the transmission power and transmission time,
respectively, $A(d)$ is the distance-dependent attenuation function, and $d$
represents the communication distance.

\textbf{Data collection latency model.} 
In each data collection cycle, the AUV visits selected relay nodes to collect data
via magnetic induction (MI) communication and subsequently forwards the aggregated
data to the base station using high-speed radio communication. Since the radio
transmission delay is negligible, the total data collection latency is defined as
\begin{equation}
T_{\text{coll}}(t) = T_{\text{comp}}(t) + T_{\text{AUV}}(t) + T_d,
\end{equation}
where $T_{\text{comp}}(t)$ denotes the computation delay at the AUV,
$T_{\text{AUV}}(t)$ is the traveling time of the AUV, and $T_d$ represents the data
uploading waiting time. In this work, $T_{\text{comp}}(t)$ required for the
AUV path planning algorithms is assumed to be negligible compared to the
traveling and data uploading delays.

\textbf{AUV mobility model.}
The AUV operates in a three-dimensional underwater environment, where its motion
is jointly affected by self-propulsion, gravity, and buoyancy. Let
$v_{\mathrm{AUV}}$ denote the magnitude of the AUV’s self-propelled velocity, and
$v_f$ denote the heave (vertical) velocity component induced by gravity and
buoyancy.
The inclination angle $\alpha$ characterizes the deviation of the AUV trajectory
from the horizontal plane and is determined by the vertical velocity component
projected along the direction toward the relay node, normalized by the AUV’s
self-propelled velocity. It is defined as
\begin{equation}
\alpha = \arcsin\left(
\frac{v_f \sin\beta}{v_{\mathrm{AUV}}}
\right),
\end{equation}
where $\beta$ is the angle between the direction toward the relaynode and the heave velocity vector.
Accordingly, the synthesized velocity of the AUV along the target direction is
given by
\begin{equation}
v_s =
\left|
\sin\!\left(
\beta + \alpha
\right)
\frac{v_{\mathrm{AUV}}}{\sin\beta}
\right|.
\end{equation}
Based on the synthesized velocity, the traveling time of the AUV along a planned
trajectory $\mathcal{H}(t)$ is computed as
\begin{equation}
T_{\mathrm{AUV}}(t)
=
\sum_{r=1}^{|\mathcal{H}(t)|}
\frac{d(\pi_{r-1},\pi_r)}{v_s(\pi_{r-1},\pi_r)},
\end{equation}
where $d(\pi_{r-1},\pi_r)$ denotes the Euclidean distance between two consecutive
waypoints $\pi_{r-1}$ and $\pi_r$.

\subsection*{Problem formulation}

Consider an IoT-enabled UWSN consisting of $N$ sensor
nodes denoted by $\mathcal{S}=\{s_1,s_2,\ldots,s_N\}$, which are randomly deployed
in a three-dimensional underwater region of interest with dimensions $L \times W \times H$. Each sensor node is aware of its three-dimensional coordinates $(x_i, y_i, z_i)$, is initialized with a limited energy budget $E_0$, has a communication radius $R_{\text{sen}}$, and remains stationary throughout the network lifetime.
 An AUV is deployed as a mobile data collector and MEC platform. In each data collection cycle, the AUV
departs from a surface base station, travels through the underwater environment to
collect data from selected relay nodes using MI communication, and
returns to the base station after completing the data collection task. The AUV mobility is governed by a realistic motion model that considers
its self-propelled velocity $v_{\text{AUV}}$ and the heave
velocity $v_f$.
 The relay nodes are dynamically selected based on the residual energy of sensor
nodes and their spatial distances, aiming to balance network energy consumption
and reduce data collection latency. Meanwhile, member nodes forward their
data to the selected relay nodes via acoustic communication. The objective of this work is to jointly design a MEC-assisted data collection
strategy and a mobility-aware AUV trajectory planning scheme that minimizes
network energy imbalance and data collection delay, while maximizing the network
lifetime under constraints on node energy budgets and AUV mobility in
IoT-enabled underwater sensor networks. The problem can be formulated as follows:\\
\textbf{Input}
\begin{itemize}
    \item A set of $N$ stationary sensor nodes $\mathcal{S} = \{ s_1, s_2, \ldots, s_N \}$, where each sensor node $s_i$ is located at position
    $\mathbf{x}_i \in \mathbb{R}^3$ and is initialized with a limited initial energy
    budget $E_0$ and a communication radius $R_{sen}$.
    \item The residual energy of sensor node $s_i$ at data collection cycle $t$,
    denoted by $E_i(t)$.
    \item Communication parameters for:
    \begin{itemize}
        \item acoustic communication among sensor nodes,
        \item magnetic induction communication between the AUV and relay nodes.
    \end{itemize}
    \item AUV mobility parameters, including the self-propelled velocity $v_{\mathrm{AUV}}$,
    the heave velocity $v_f$, and the synthesized velocity $v_s$, which jointly
    determine the AUV traveling time $T_{\mathrm{AUV}}(t)$.

    \item The energy consumption and data collection latency models introduced in
    the Preliminaries section.
\end{itemize}

\textbf{Output}
\begin{itemize}
    \item Relay node selection $\mathcal{S}_{\mathrm{relay}}(t) \subseteq \mathcal{S},$ where $\mathcal{S}_{\mathrm{relay}}(t)$ denotes the set of selected relay nodes visited by the AUV in cycle $t$.
    \item AUV trajectory planning
       $ \mathcal{H}(t) = \{ \pi_0, \pi_1, \ldots, \pi_{|\mathcal{H}(t)|} \}$,
    where $\pi_0$ and $\pi_{|\mathcal{H}(t)|}$ represent the base station location, and the intermediate waypoints correspond to the selected relay nodes.
\end{itemize}

\textbf{Objective}
The objective is to jointly optimize relay node selection and AUV trajectory
planning in order to:
\begin{itemize}
    \item minimize data collection delay,
    \item maximize the network lifetime.
\end{itemize}

\textbf{Constraints}
The joint optimization of relay node selection and AUV trajectory planning is
subject to the following constraints.

\begin{align}
    & E_i(t) \ge 0, \quad \forall s_i \in \mathcal{S}, \; \forall t, 
    \label{c:energy} \\[2pt]
    & s_i \in \mathcal{S}_{\mathrm{relay}}(t) \Rightarrow E_i(t) \ge E_{\min},
    \label{c:target} \\[2pt]
    & \pi_0 = \pi_{|\mathcal{H}(t)|} = \text{base station}.
    \label{c:trajectory}
\end{align}

\section{Proposed algorithm}
To solve MECEBDC, it is to enable energy-efficient data collection in underwater sensor networks with AUV-assisted mobile edge computing, this paper presents a two-stage solution framework. First, the network is partitioned into 
$C$ clusters using an enhanced K-means algorithm that jointly accounts for spatial distance, residual node energy, and intra-cluster communication range, yielding a set of cluster heads and cluster structures. Subsequently, metaheuristic optimization algorithms, including Genetic Algorithm (GA) and Particle Swarm Optimization (PSO), are utilized to determine an efficient AUV trajectory for visiting the selected cluster heads while minimizing the overall travel time. These algorithms have proven highly effective in addressing optimization problems in stochastic and uncertain environments \cite{binh2023efficient}. Compared with greedy strategies used in prior work, the proposed approach achieves near-global optimality, thereby reducing data collection latency and prolonging network lifetime. The scheme proposed is  represents the more detail in Algorithm \ref{alg:framework}.

\begin{algorithm}[H]
\begin{minipage}{0.8\linewidth}
\footnotesize
\caption{Cyclic AUV-Assisted Data Collection Procedure}
\label{alg:framework}
\begin{algorithmic}[1]

\Require Sensor node set $V$ with initial energy $E_i$
\Ensure Network lifetime $T$

\State Perform constrained K-means clustering
\State Select one Relay Node (RN) per cluster based on residual energy
\State Compute AUV visiting sequence using metaheuristics path planning algorithm

\State $t \gets 0$
\While{alive node ratio $> 10\%$}

    \State Execute data collection
    \State Update residual energy of all nodes
    \State Remove depleted nodes
    
    \If{node failures or topology changes occur}
        \State Re-cluster alive nodes
        \State Recompute AUV route using a heuristic algorithm
    \EndIf
    
    \State $t \gets t + 1$

\EndWhile
\State \Return $t$
\end{algorithmic}
\end{minipage}
\end{algorithm}

\textbf{Individual representation.} 
In both GA and PSO, the solution representation for the MECEBDC problem is identical. Each individual is encoded as a permutation of relay node indices, representing the visiting sequence of the AUV. Let
$C$ denote the number of relay nodes; a feasible solution is thus expressed as
\begin{equation}
    \mathcal{r} = [0, i_1, i_2, \dots, i_C],
\end{equation}
where index $0$ is the base station and $\{i_1, i_2, \dots, i_C\}$ is a permutation of $\{1,2,\dots,C\}$. Each node appears exactly once, ensuring that all relay nodes are visited without repetition.
The fitness of a solution is defined as the total traveling time:
\begin{equation}
f(\mathcal{r}) = T_{AUV}(\mathcal{r}) =
 \sum_{i=0}^{C-1} T_{\mathcal{r}(i),\mathcal{r}(i+1)} + T_{\mathcal{r}(C),\mathcal{r}(0)},
\end{equation}
where $T_{ij}$ denotes the travel time between nodes $i$ and $j$. Since the objective is to minimize travel time, solutions with smaller $f(\mathcal{r})$ correspond to higher-quality routes.
For population-based algorithms (GA and PSO variants), a shared initialization strategy is adopted to ensure fair comparison. An initial population (or swarm) of size $P$ is generated by randomly permuting the Relay Node indices while fixing the base station at the first position. Each individual in the population therefore satisfies the feasibility constraints by construction.

\subsection{Genetic algorithm}
The Genetic Algorithm is a population-based evolutionary optimization method inspired by natural selection. It iteratively improves candidate solutions through selection, crossover, mutation, and elitism to balance exploration and exploitation.

\begin{algorithm}[H]
\caption{Genetic Algorithm (GA) for AUV Routing}
\label{alg:ga_core}
\begin{algorithmic}[1]

\Require The population $P$, generations $T$, rates $p_c, p_m$, elitism size $k_e$
\Ensure Best solution $X^\star$
\State Initialize population $\mathcal{P} = \{\nabla_i\}_{i=1}^P$, where $\nabla_i = [0, i_1, i_2, \dots, i_C]$
 (valid permutations; depot fixed at 0)
\State Evaluate objective $J(X)$ for all $X\in\mathcal{P}$; set $X^\star \gets \arg\min_{X\in\mathcal{P}} J(X)$

\For{$t=1$ to $T$}
    \State $\mathcal{E} \gets$ best $k_e$ individuals in $\mathcal{P}$ 
    \State $\mathcal{P}' \gets \mathcal{E}$
    \While{$|\mathcal{P}'| < P$}
        \State Select parents $(p_1,p_2)$ (tournament selection)
        \State Create offspring via OX crossover with prob. $p_c$; otherwise copy parents
        \State Apply inversion mutation to offspring with prob. $p_m$
        \State Add offspring to $\mathcal{P}'$
    \EndWhile
    \State $\mathcal{P} \gets \mathcal{P}'$; evaluate $J(\cdot)$; update $X^\star$ if improved
\EndFor
\State \Return $X^\star$
\end{algorithmic}
\end{algorithm}

\subsection{Particle Swarm Optimization}
Particle Swarm Optimization is a swarm-based metaheuristic where individuals update their positions according to inertia, personal experience, and social interaction. The collective learning mechanism enables efficient search through cooperation among particles.

\begin{algorithm}[H]
\caption{Swap-based PSO for AUV Routing}
\label{alg:pso_routing}
\begin{algorithmic}[1]

\Require Relay node coordinates $coords$, particle number $P$, iteration number $T$, parameters $c_1, c_2$
\Ensure Best route $Gbest$ and best travel time $Gcost$

\State Initialize particles $X_i= \nabla_i$ where $\nabla_i = [0, i_1, i_2, \dots, i_C]$ as permutations (depot fixed at 0)
\State Initialize velocities $V_i \gets \emptyset$
\State Evaluate $J_i = \text{TravelTime}(X_i)$
\State Set $Pbest_i \gets X_i$; $m \gets \arg\min_{1 \le i \le P} J_i$
\State $Gbest\gets X_{m}$; $Gcost\gets J_{m}$

\For{$t = 1$ to $T$}
    \State $\alpha \gets 0.7 - 0.5 \cdot (t/T)$
    \For{each particle $i$}
        \State $V_i \gets \alpha V_i \oplus c_1\,\text{Swap}(X_i,Pbest_i) \oplus c_2\,\text{Swap}(X_i,Gbest)$
        \State $X_i \gets \text{Apply}(V_i, X_i)$
        \State Evaluate $J_i = \text{TravelTime}(X_i)$
        \State Update $Pbest_i$ and $Gbest$ if improved
    \EndFor
\EndFor

\State \Return $Gbest$, $Gcost$

\end{algorithmic}
\end{algorithm}

\subsection{Adaptive-Enhanced PSO (AE-PSO)}
Adaptive-Noise PSO extends the standard PSO by incorporating a dynamically adjusted stochastic perturbation mechanism. The noise intensity increases under stagnation to enhance exploration and decreases otherwise to maintain convergence stability.Compared with standard PSO, AE-PSO incorporates controlled diversification to alleviate premature convergence without sacrificing convergence stability. By adaptively adjusting perturbation intensity, the algorithm maintains a balanced exploration–exploitation trade-off throughout the search process.

\begin{algorithm}[H]
\caption{Swap-based AE-PSO for AUV Routing}
\label{alg:pso_adaptive_noise_short}
\begin{algorithmic}[1]

\Require Relay node coordinates $coords$, particle number $P$, iteration number $T$, parameters $c_1, c_2$
\Ensure Best route $Gbest$ and best travel time $Gcost$

\State Initialize particles $X_i= \nabla_i$ where $\nabla_i = [0, i_1, i_2, \dots, i_C]$ as permutations (depot fixed at 0)
\State Initialize velocities $V_i \gets \emptyset$
\State Evaluate $J_i = \text{TravelTime}(X_i)$
\State Set $Pbest_i \gets X_i$; $m \gets \arg\min_{1 \le i \le P} J_i$
\State $Gbest\gets X_{m}$; $Gcost\gets J_{m}$
\State Initialize noise control: $p\in[p_{\min},p_{\max}]$, $s\gets0$, $J^\star\gets Gcost$

\For{$t=1$ to $T$}
  \State $\alpha\gets 0.7-0.5(t/T)$
  \For{$i=1$ to $P$}
    \State $V_i\gets \alpha V_i \oplus c_1\,Swap(X_i,Pbest_i)\oplus c_2\,Swap(X_i,Gbest)$
    \State With prob. $p$: $V_i\gets V_i\oplus RandSwap(\le2)$
    \State $X_i\gets Apply(V_i,X_i)$; $J_i\gets TravelTime(X_i)$; update $Pbest_i$ and $(Gbest,Gcost)$ if improved
  \EndFor
  \State $s \gets (Gcost \ge J^\star-\varepsilon)?(s+1):0$; if $s=0$ then $J^\star\gets Gcost$
  \State $p \gets (s>15)?\min(p_{\max},1.3p):\max(p_{\min},0.95p)$
\EndFor

\State \Return $Gbest$, $Gcost$
\end{algorithmic}
\end{algorithm}

\subsection{Lévy Flight PSO (LF-PSO)}
Lévy-Flight PSO integrates Lévy-distributed perturbations into the update process to enable occasional long-range transitions. The heavy-tailed behavior of Lévy flights enhances global exploration and reduces the risk of premature convergence.

Compared with AE-PSO, LF-PSO provides significantly stronger exploration capability. Whereas AE-PSO relies on adaptively controlled random perturbations, LF-PSO enables more aggressive diversification through heavy-tailed step lengths, thereby improving the ability to escape deep local optima in complex search landscapes. However, such strong perturbations may reduce convergence stability relative to AE-PSO. Consequently, LF-PSO is particularly suitable for large-scale or highly multimodal problems requiring enhanced global exploration, while AE-PSO remains preferable for moderately complex search spaces where stable convergence is desired.

\begin{algorithm}[H]
\caption{Swap-based LF-PSO for AUV Routing}
\label{alg:pso_levy}
\begin{algorithmic}[1]

\Require Relay node coordinates $coords$, particle number $P$, iteration number $T$, parameters $c_1, c_2$
\Ensure Best route $Gbest$ and best travel time $Gcost$

\State Initialize particles $X_i= \nabla_i$ where $\nabla_i = [0, i_1, i_2, \dots, i_C]$ as permutations (depot fixed at 0)
\State Initialize velocities $V_i \gets \emptyset$
\State Evaluate $J_i = \text{TravelTime}(X_i)$
\State Set $Pbest_i \gets X_i$; $m \gets \arg\min_{1 \le i \le P} J_i$
\State $Gbest\gets X_{m}$; $Gcost\gets J_{m}$
\State Set Lévy parameters: $p_{\ell}\gets 0.15$, $\beta\gets 1.5$, $k_{\max}\gets 4$

\For{$t = 1$ to $T$}
    \State $\alpha \gets 0.7 - 0.5 \cdot (t/T)$
    \For{each particle $i$}
        \State $V_i \gets \alpha V_i \oplus c_1\,\text{Swap}(X_i,Pbest_i) \oplus c_2\,\text{Swap}(X_i,Gbest)$
        \If{$rand() < p_{\ell}$}
            \State $k \gets \min(k_{\max},\,\max(1,\lfloor \text{LevyStep}(\beta)\rfloor))$
            \State $V_i \gets V_i \oplus \text{RandSwap}(k)$
        \EndIf
        \State $X_i \gets \text{Apply}(V_i, X_i)$; \quad $J_i \gets \text{TravelTime}(X_i)$
        \State Update $Pbest_i$ and $(Gbest,Gcost)$ if improved
    \EndFor
\EndFor

\State \Return $Gbest$, $Gcost$

\end{algorithmic}
\end{algorithm}

\section{Experimental settings and results analysis}\label{sec:results}
\subsection{Experimental settings}
Simulation experiments are conducted using a Python-based implementation executed on the online Kaggle platform.
A total of 150 to 550 sensor nodes are randomly deployed in a three-dimensional underwater monitoring area of $400\,\mathrm{m} \times 400\,\mathrm{m} \times 400\,\mathrm{m}$.
An autonomous underwater vehicle (AUV) is employed to collect data from the network in a cyclic manner.
The acoustic transmission rate is set to 4 kbps, and the AUV moves with a constant cruising speed while taking into account the influence of underwater currents.
The detailed simulation parameters are summarized in Table~\ref{tab:simulation_parameters}.

The simulation performance is evaluated using several metrics, including network lifetime, AUV traveling time, and the standard deviation (STD) of travel time across five different algorithms.
Network lifetime is defined as the total number of data collection cycles completed before at least 90\% of the sensor nodes deplete their energy, i.e., when the ratio of alive nodes drops below 10\%.
These metrics are used to comprehensively assess the efficiency, stability, and scalability of the proposed algorithms.

\begin{table}[H]
\caption{Simulation parameters}
\label{tab:simulation_parameters}
\centering
\begin{tabular}{l c}
\hline
\textbf{Parameter} & \textbf{Value} \\
\hline
Network size $(L \times W \times H)$      
& $400\,\mathrm{m} \times 400\,\mathrm{m} \times 400\,\mathrm{m}$ \\

Number of sensor nodes $N$            
& $150$--$550$ \\

Sensing range $(R_{sen})$                 
& $50$--$60\,\mathrm{m}$ \\

AUV self speed $(v_{\mathrm{AUV}})$            
& $3\,\mathrm{m/s}$ \\

AUV heave velocity $(v_f)$                
& $1.2\,\mathrm{m/s}$ \\

Packet size $(l)$                         
& $1024\,\mathrm{bit}$ \\

Data rate for acoustic communication $(DR)$ 
& $4\,\mathrm{kbps}$ \\

Data rate for MI communication $(DR_i)$   
& $10\,\mathrm{kbps}$ \\

Packet generation rate per sensor         
& $100$ packets per round \\

Initial node energy $(E_0)$               
& $100\,\mathrm{J}$ \\

Data collection power consumption $(P_c)$            
& $0.8 \times 10^{-3}\,\mathrm{W}$ \\

Transmission power consumption $(P_t)$       
& $1.6 \times 10^{-3}\,\mathrm{W}$ \\

\hline
\end{tabular}
\end{table}

\subsection{Experimental results analysis}
\subsubsection{Convergence analysis}
\begin{figure}[H]
    \centering

    \begin{subfigure}{0.48\linewidth}
        \centering
        \includegraphics[width=\linewidth]{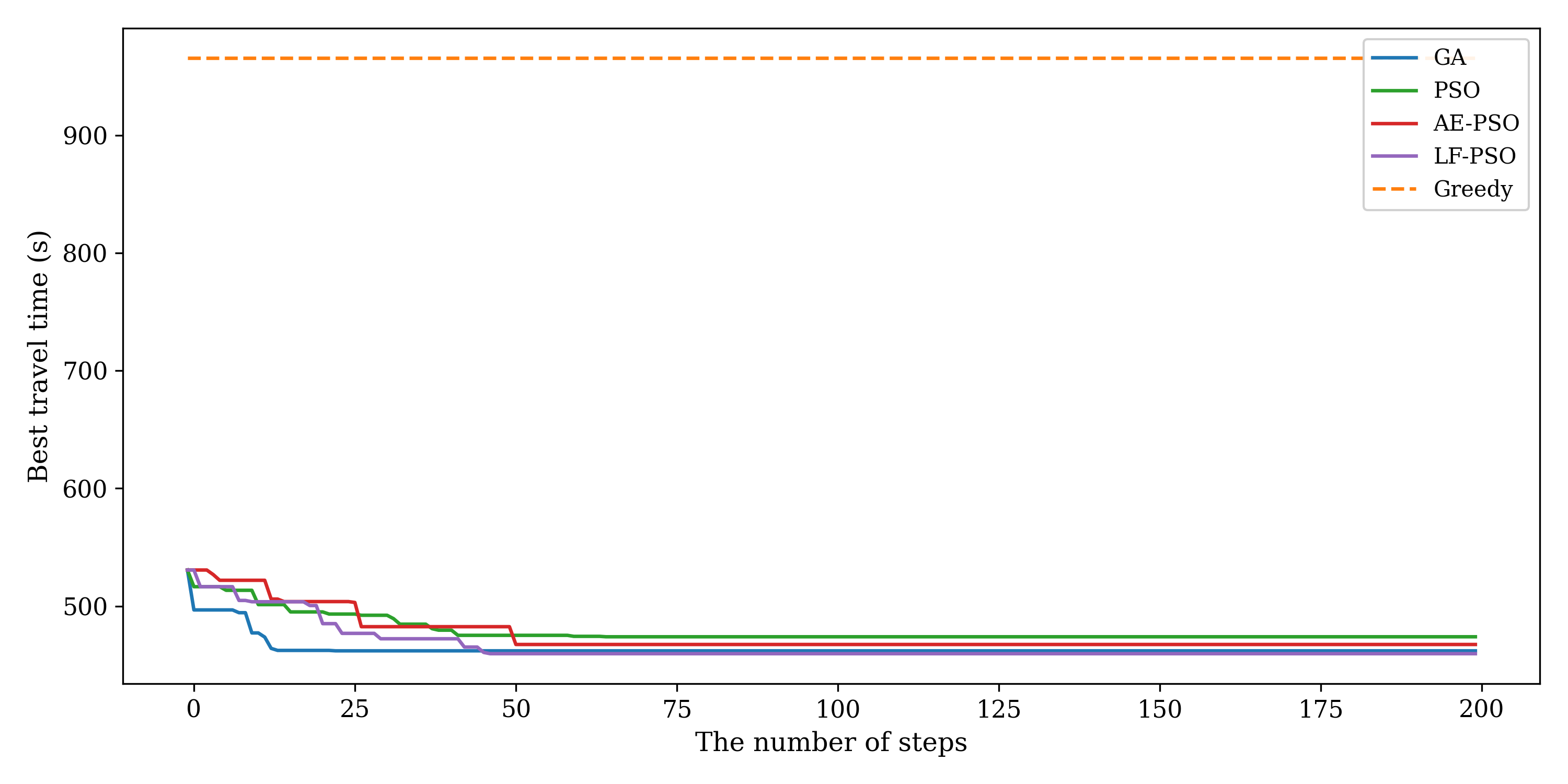}
        \caption{Algorithm convergence}
        \label{fig:conv_150}
    \end{subfigure}
    \hfill
    \begin{subfigure}{0.48\linewidth}
        \centering
        \includegraphics[width=\linewidth]{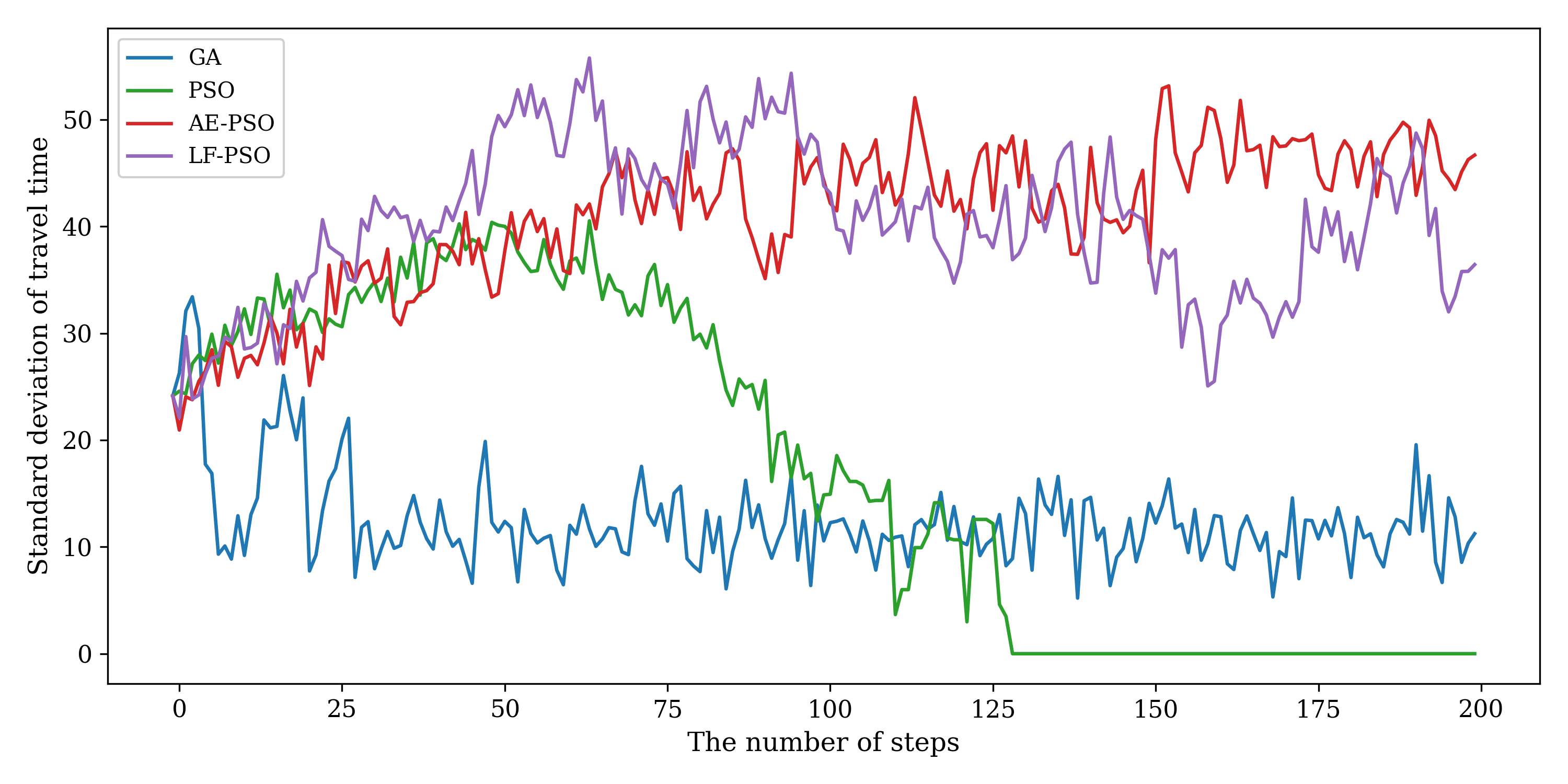}
        \caption{Traveling time standard deviation}
        \label{fig:std_150}
    \end{subfigure}

    \caption{Convergence behavior and stability analysis of different algorithms in a 150-node network}
    \label{fig:conv_std_150}
\end{figure}

Figure \ref{fig:conv_150} compares the convergence behaviors of different algorithms for a 150-node network. Metaheuristic methods rapidly reduce travel time in early iterations, while the Greedy baseline \cite{cai2019data} shows no improvement. GA converges fastest but stabilizes early, whereas PSO variants, especially AE-PSO and LF-PSO, achieve better final solutions through sustained exploration, with LF-PSO yielding the best result in this run.

To further interpret the convergence behavior observed in Fig.\ref{fig:conv_150}, Fig.\ref{fig:std_150} presents the standard deviation of travel time over optimization steps, providing insights into the stability and robustness of the considered algorithms. GA consistently maintains the lowest standard deviation throughout the optimization process, indicating highly stable convergence with limited fluctuations. The standard PSO initially exhibits moderate variability, which gradually decreases as the optimization progresses. In contrast, AE-PSO and LF-PSO demonstrate higher variability during the early and middle stages of optimization, reflecting stronger exploration behavior. LF-PSO preserves a relatively high level of variability throughout the optimization process, indicating sustained exploration enabled by Lévy-flight perturbations.

\subsubsection{Impact of Network Size on System Performance}
\begin{figure}[H]
    \centering

    \begin{subfigure}{0.48\linewidth}
        \centering
        \includegraphics[width=\linewidth]{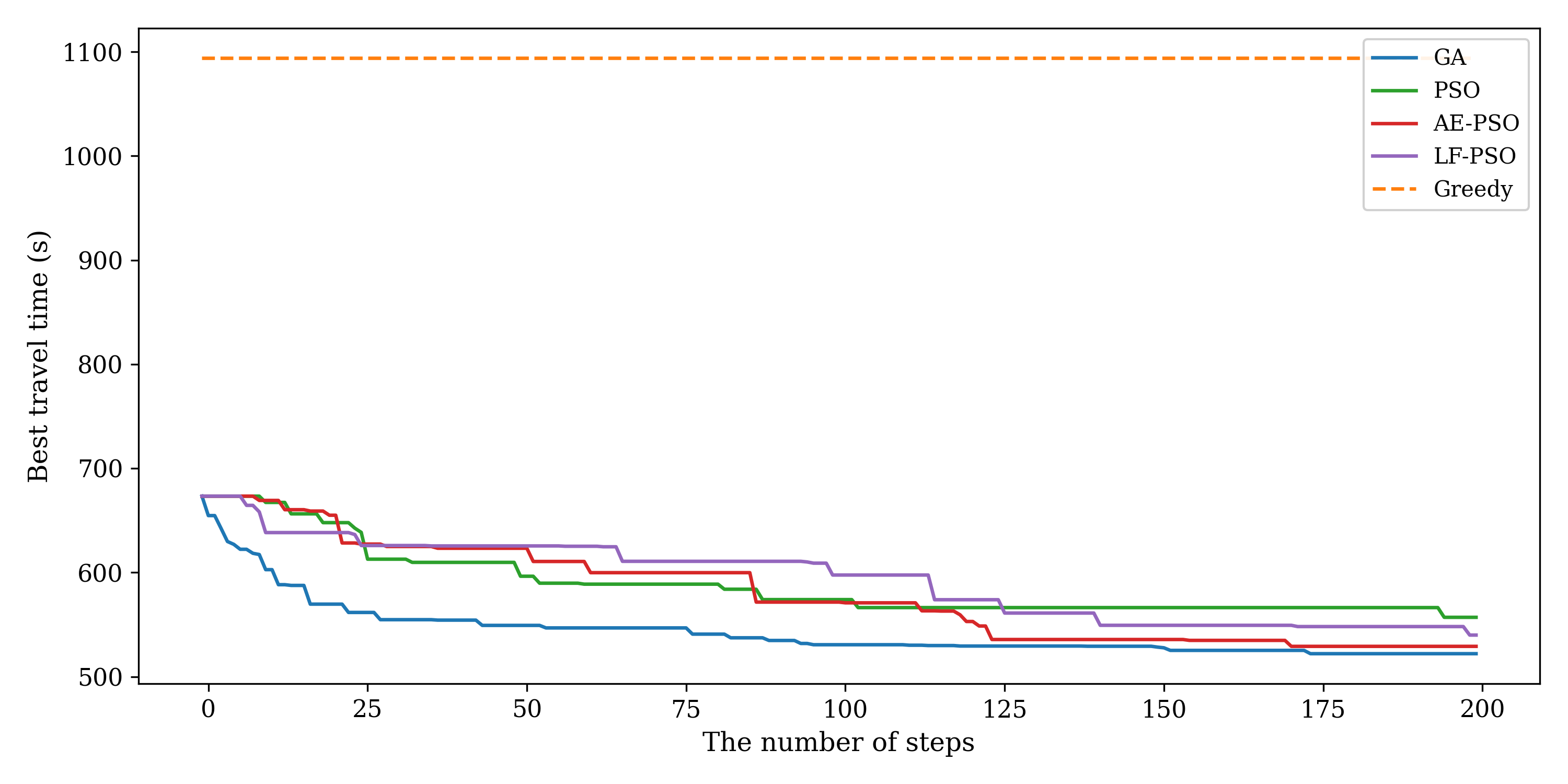}
        \caption{Algorithm convergence}
        \label{fig:conv_350}
    \end{subfigure}
    \hfill
    \begin{subfigure}{0.48\linewidth}
        \centering
        \includegraphics[width=\linewidth]{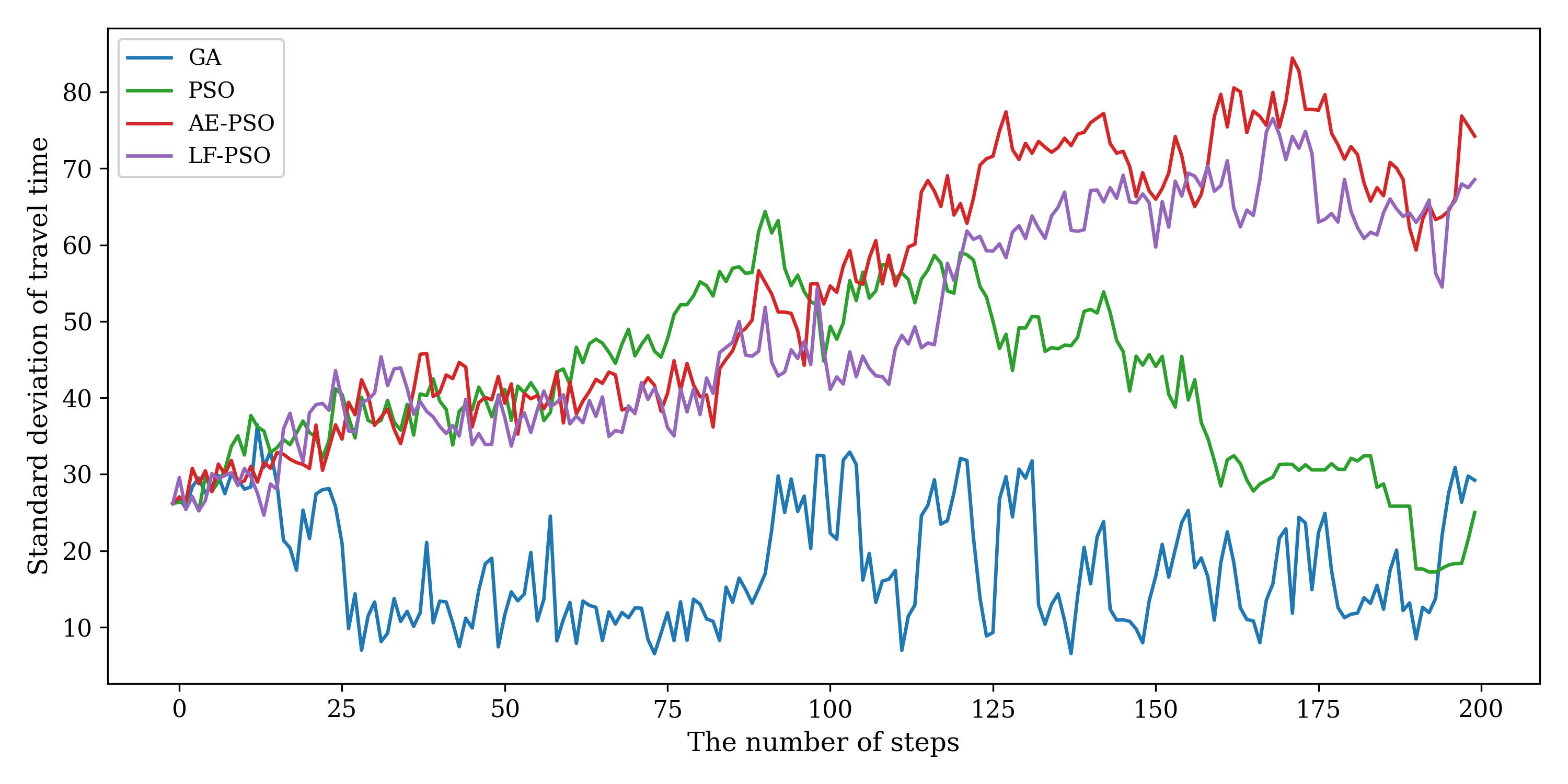}
        \caption{Traveling time standard deviation}
        \label{fig:std_350}
    \end{subfigure}

    \caption{Convergence behavior and stability analysis of different algorithms in a 350-node network}
    \label{fig:conv_std_350}
\end{figure}

\begin{figure}[H]
    \centering

    \begin{subfigure}{0.48\linewidth}
        \centering
        \includegraphics[width=\linewidth]{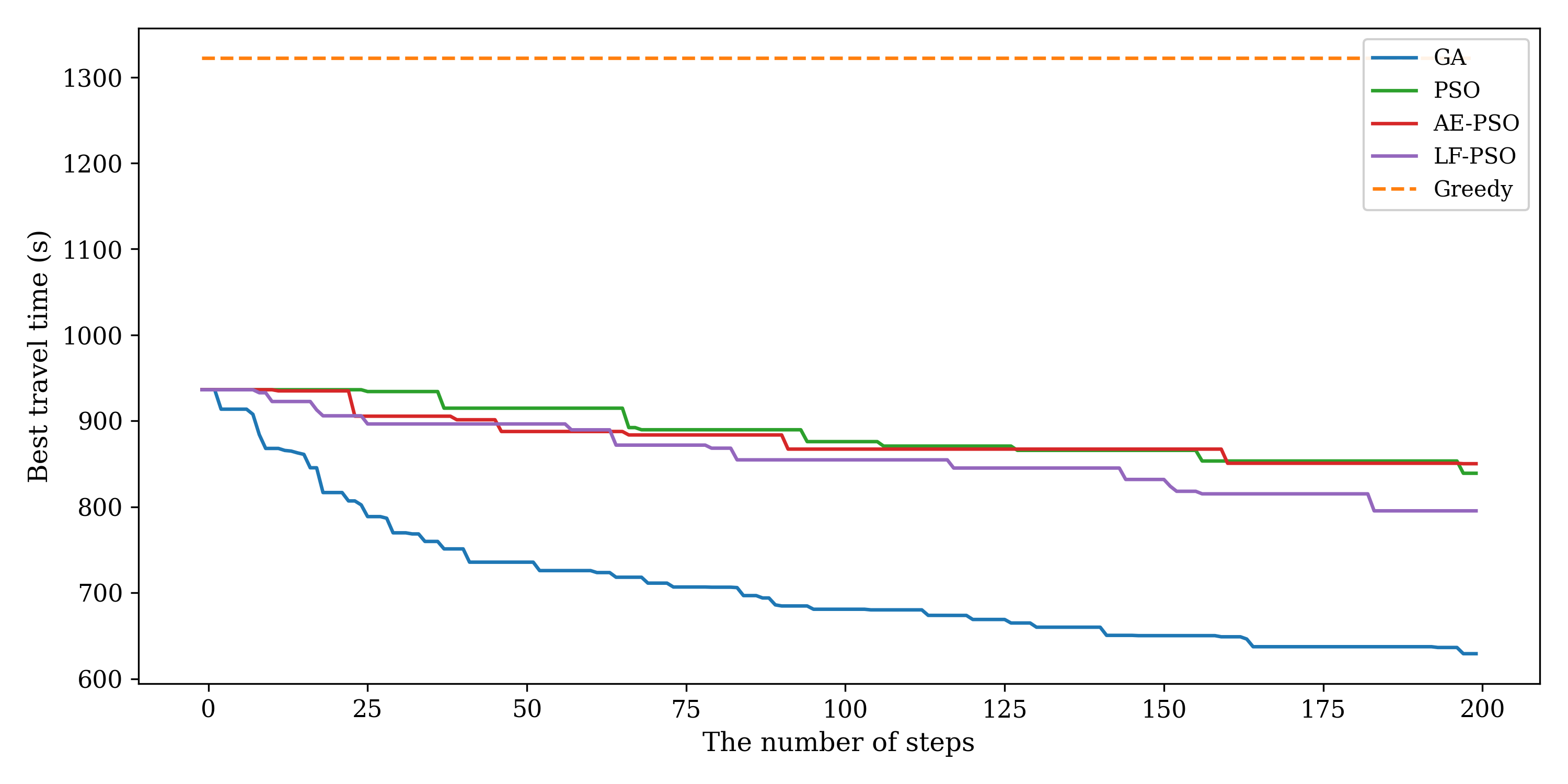}
        \caption{Algorithm convergence}
        \label{fig:conv_550}
    \end{subfigure}
    \hfill
    \begin{subfigure}{0.48\linewidth}
        \centering
        \includegraphics[width=\linewidth]{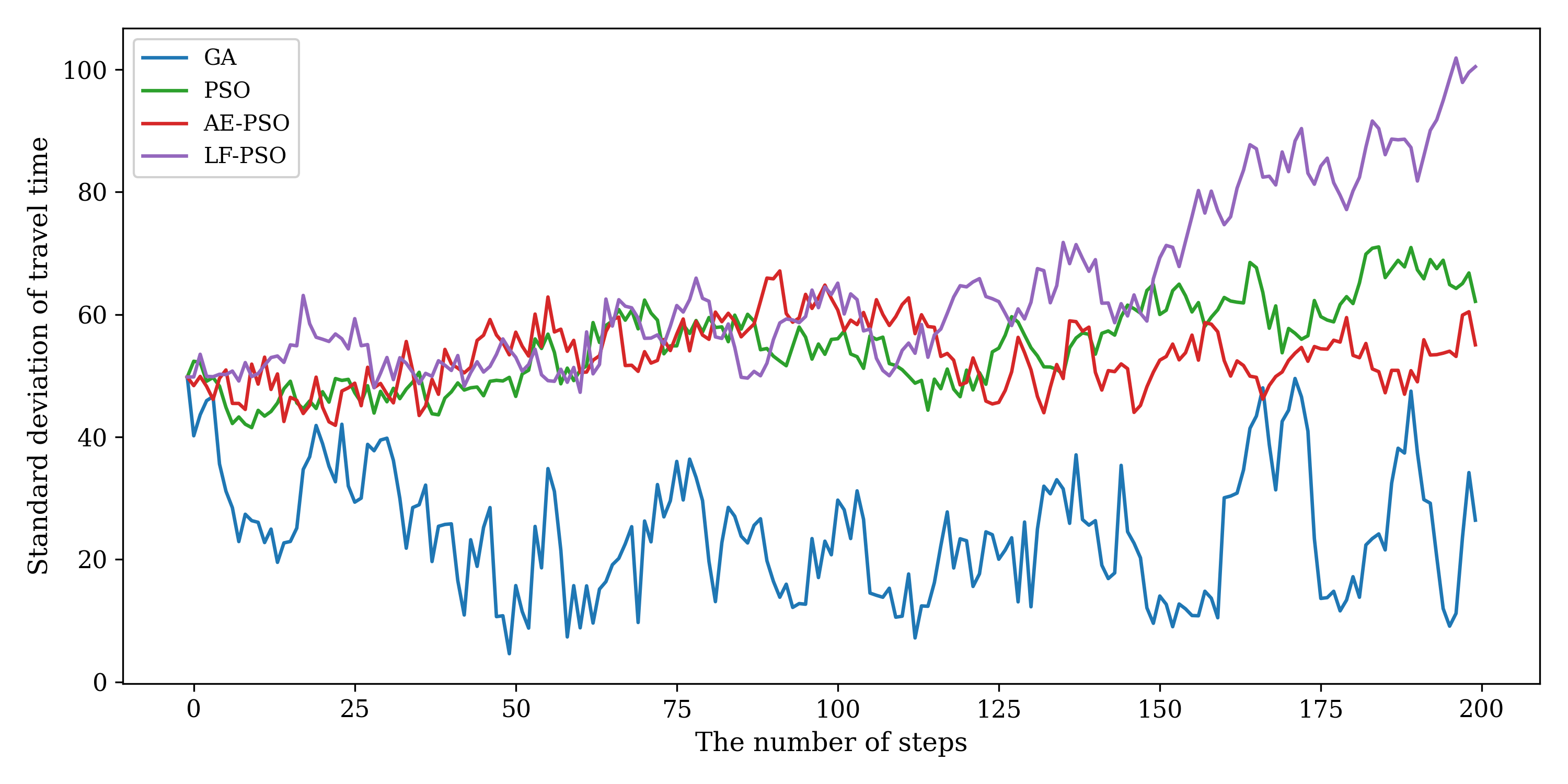}
        \caption{Traveling time standard deviation}
        \label{fig:std_550}
    \end{subfigure}
    \caption{Convergence behavior and stability analysis of different algorithms in a 550-node network}
    \label{fig:conv_std_550}
\end{figure}

\textit{a. Impact on algorithm behavior}\\
This subsection investigates the impact of network size on algorithm performance in terms of convergence behavior and standard deviation, considering medium-scale (350 nodes) and large-scale (550 nodes) scenarios shown in Figs.~\ref{fig:conv_std_350} and \ref{fig:conv_std_550}. In Fig.~\ref{fig:conv_std_350}, standard PSO exhibits low variance and rapid stabilization, indicating limited exploration. As network size increases (Fig.~\ref{fig:conv_std_550}), PSO variants maintain higher variability to cope with increased search complexity. In contrast, GA preserves stable diversity across both scales, achieving consistently superior solution quality. LF-PSO further demonstrates strong exploration capability in large-scale networks.\\

\noindent\textit{b. Impact on network lifetime}\\

\begin{figure}[H]
    \centering
    \includegraphics[width=0.7\linewidth]{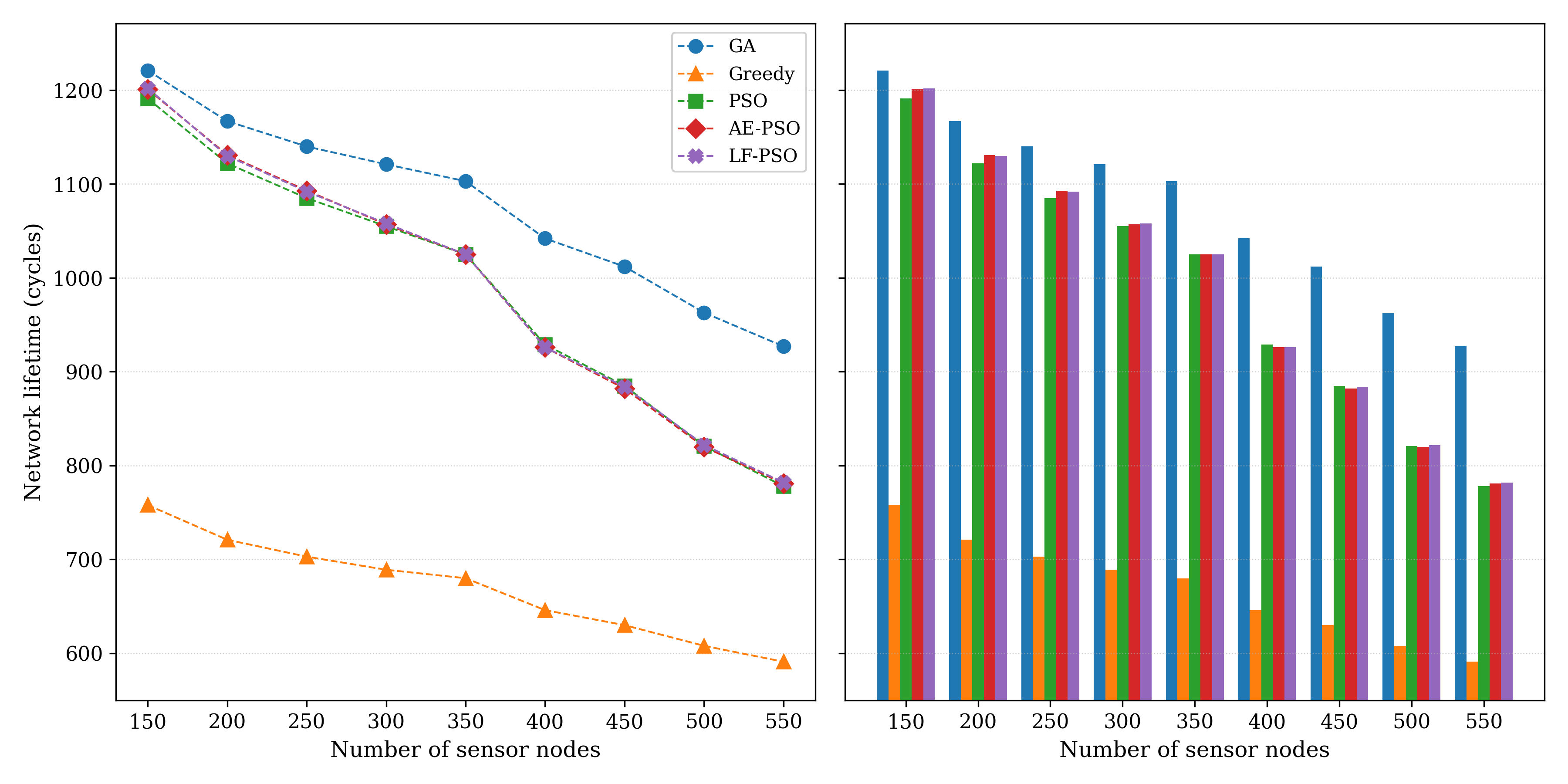}
    \caption{Network lifetime under different network sizes}
    \label{fig:lifetime}
\end{figure}

This subsection analyzes the impact of network size on network lifetime under different routing strategies. As shown in Fig.~\ref{fig:lifetime}, the lifetime of the network consistently decreases as the number of sensor nodes increases for all algorithms evaluated. This decline is primarily attributed to higher communication overhead and longer AUV travel paths in denser networks, which increase energy consumption per data collection cycle and accelerate node depletion. The combined effect of increased relay traffic, longer visit times, and more frequent transmissions further intensifies the energy imbalance between nodes. Among the compared methods, GA achieves the longest network lifetime in all network sizes. Its superior global optimization capability produces more efficient routes, reducing travel time and distributing energy consumption more evenly, thereby delaying the exhaustion of critical nodes. The PSO-based variants exhibit comparable performance, with AE-PSO and LF-PSO slightly outperforming standard PSO due to their enhanced exploration mechanisms that mitigate premature convergence. In contrast, the Greedy method consistently results in the shortest lifetime, as its lack of global optimization leads to suboptimal routing and faster energy depletion, particularly in large-scale networks. Therefore, while increasing network size inherently reduces lifetime, effective global optimization can significantly mitigate this degradation.

\section{Conclusion}
This paper presented a mobility-aware mobile edge computing framework for energy-balanced data collection in IoT-enabled underwater sensor networks. By jointly optimizing energy-aware clustering, relay node selection, and AUV trajectory planning under realistic underwater mobility constraints, the proposed approach effectively addresses energy imbalance and data collection inefficiency in UWSNs. Extensive simulations evaluated convergence behavior, stability, scalability, and network lifetime across different network sizes. The results show that the genetic algorithm consistently achieves superior performance in terms of travel time minimization and network lifetime extension, owing to its effective balance between exploration and exploitation. While standard PSO suffers from premature convergence, advanced PSO variants improve robustness by maintaining population diversity, with LF-PSO demonstrating competitive performance in large-scale networks. Moreover, the results confirm that increasing network density significantly impacts network lifetime, highlighting scalability challenges in underwater IoT systems. Overall, the proposed framework provides an effective and scalable solution for long-term AUV-assisted data collection in energy-constrained underwater sensor networks.
\nocite{*}
\bibliographystyle{splncs04} 
\bibliography{reference}
\end{document}